\documentclass{article}

\usepackage{arxiv}

\usepackage{cite}
\usepackage[utf8]{inputenc} % allow utf-8 input
\usepackage[T1]{fontenc}    % use 8-bit T1 fonts
\usepackage{hyperref}       % hyperlinks
\usepackage{url}            % simple URL typesetting
\usepackage{booktabs}       % professional-quality tables
\usepackage{amsfonts}       % blackboard math symbols
\usepackage{nicefrac}       % compact symbols for 1/2, etc.
\usepackage{microtype}      % microtypography
\usepackage{lipsum}		% Can be removed after putting your text content
\usepackage{graphicx}
\usepackage{doi}
\usepackage{amsmath}
\usepackage{pgfplots, tikz}
\usetikzlibrary{shapes,arrows}
\usetikzlibrary{arrows,positioning,shapes.geometric}
\usepackage{makecell}
\usepackage{algorithm}
\usepackage{algorithmic}
\usepackage{multirow}
\usepackage{listings}

\usepackage[square,sort,comma,numbers]{natbib}
\title{Disentangling Homophemes in Lip Reading using Perplexity Analysis}

\author{ \href{https://orcid.org/0000-0000-0000-0000}{\includegraphics[scale=0.06]{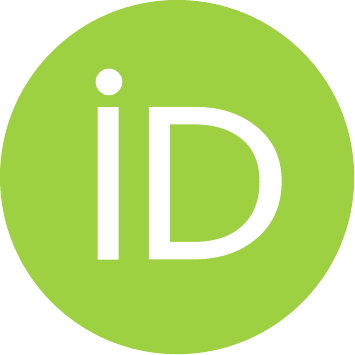}\hspace{1mm}Souheil Fenghour}\thanks{Use footnote for providing further
		information about author (webpage, alternative
		address)---\emph{not} for acknowledging funding agencies.} \\
	Department of Engineering\\
	London South Bank University\\
	London, United Kingdom \\
	\texttt{fenghous@lsbu.ac.uk} \\
	\And
	\href{https://orcid.org/0000-0000-0000-0000}{\includegraphics[scale=0.06]{orcid.pdf}\hspace{1mm}Daqing Chen} \\
	Department of Engineering\\
	London South Bank University\\
	London, United Kingdom \\
	\texttt{chend@lsbu.ac.uk} \\
	\And
	\href{https://orcid.org/0000-0000-0000-0000}{\includegraphics[scale=0.06]{orcid.pdf}\hspace{1mm}
 	Perry Xiao} \\
  	Department of Engineering\\
  	London South Bank University,\\
 	London, United Kingdom \\
  	\texttt{perry.xiao@lsbu.ac.uk} \\
  	\And
	\href{https://orcid.org/0000-0000-0000-0000}{\includegraphics[scale=0.06]{orcid.pdf}\hspace{1mm}
 	Kun Guo} \\
  	Xi'an VANXUM Electronics Technology Co. Ltd.\\
 	Xi'an, China\\
  	\texttt{guokun84@126.com} \\
}

\renewcommand{\shorttitle}{\textit{arXiv} Template}

\hypersetup{
pdftitle={A template for the arxiv style},
pdfsubject={q-bio.NC, q-bio.QM},
pdfauthor={Souheil Fenghour, Daqing Chen, Kun Guou, Perry Xiao},
pdfkeywords={First keyword, Second keyword, More},
}

\begin{document}

% Define block styles
\tikzstyle{block} = [rectangle, draw, fill=blue!20, text width=8em, text centered, minimum height=4em]
\tikzstyle{decision} = [diamond, draw, fill=blue!20, text width=5.6em, text centered, node distance=3.6cm, inner sep=0pt]
\tikzstyle{empty} = [text width=2em, text centered, minimum height=4em]
\tikzstyle{line} = [draw, -latex']

\tikzstyle{empty3} = [rectangle, text width=5em, text centered, minimum height=4em]
\tikzstyle{block3} = [rectangle, draw, fill=blue!20, text width=4em, text centered, minimum height=2em, minimum width=4em]

\tikzstyle{startstop} = [rectangle, rounded corners, minimum width=3cm, minimum height=1cm,text centered, draw=black, fill=red!30]
\tikzstyle{io} = [trapezium, trapezium left angle=70, trapezium right angle=-70, inner sep=8pt, outer sep=0pt, minimum width=3cm, minimum height=1cm, text centered, draw=black, fill=blue!30]
\tikzstyle{process} = [rectangle, minimum width=3cm, minimum height=1cm, text centered, draw=black, fill=orange!30]
\tikzstyle{decision} = [diamond, minimum width=3cm, minimum height=1cm, text centered, draw=black, fill=green!30]
\tikzstyle{arrow} = [->,>=stealth]

\maketitle

\begin{abstract}
The performance of automated lip reading using visemes as a classification schema has achieved less success compared with the use of ASCII characters and words largely due to the problem of different words sharing identical visemes. The Generative Pre-Training transformer is an effective autoregressive language model used for many tasks in Natural Language Processing, including sentence prediction and text classification. This paper proposes a new application for this model and applies it in the context of lip reading, where it serves as a language model to convert visual speech in the form of visemes, to language in the form of words and sentences. The network uses the search for optimal perplexity to perform the viseme-to-word mapping and is thus a solution to the one-to-many mapping problem that exists whereby various words that sound different when spoken look identical.

This paper proposes a method to tackle the one-to-many mapping problem when performing automated lip reading using solely visual cues in two separate scenarios: the first scenario is where the word boundary, that is, the beginning and the ending of a word, is unknown; and the second scenario is where the boundary is known.

Sentences from the benchmark BBC dataset "Lip Reading Sentences in the Wild"(LRS2), are classified with a character error rate of 10.7\% and a word error rate of 18.0\%. The main contribution of this paper is to propose a method of predicting words through the use of perplexity analysis when only visual cues are present, using an autoregressive language model. 

\end{abstract}

% keywords can be removed

\section{Introduction}
\label{sec:Introduction}

The task of automated lip reading has attracted a lot of research attention in recent years and has made significant breakthrough with a variety of machine learning based approaches having been implemented \cite{b1}\cite{b2}. Automated lip reading can be done both with and without the assistance of audio \cite{b3}, When performed without the presence of audio, it is often referred to as visual speech recognition \cite{b4}. Though the majority of approaches to automated lip reading have largely focused on decoding long speech segments in the form of words and sentences using either words ASCII characters as the class, visemes are an alternative classification schema with several advantages over words and ASCII symbols \cite{b5}\cite{b6}\cite{b7}\cite{b8}\cite{b9}\cite{b10}.

Visemes represent images of lip movements and are more directly correlated with visual speech. In comparison to the use of either words or ASCII characters as classes, the number of classes that is needed to classify visemes is smaller and this helps to reduces computational bottleneck. Furthermore, a viseme-based classification schema does not require pre-trained lexicons, meaning that a viseme-based lip reading system can be used to classify words that have not presented in the training phase. A viseme-based lip reading system can even be generalised to different languages because many languages share identical visemes.

There are limitations to the use of visemes in automated lip reading. Unlike ASCII where there is a straightforward conversion of a set of recognised ASCII characters to possible words in a one-to-one mapping relationship, for classifying visemes, this one-to-one mapping relationship does not always exist as one set of visemes can map to multiple different sounds or phonemes. There are many homopheme words, i.e., words that look the same when spoken but sound different \cite{b11}, as such there exists a one-to-many mapping relationship between visemes (distinct visual units of speech), and phonemes (distinct acoustic units of speech). If the words uttered by an individual speaking were to be decoded using only visual information, it would in many cases be impossible to always narrow down what has been said to only one possible set of words, phrases or sentences.
 
Alternatively, speech can be classified in the form of phonemes as there are far fewer homophonic words in the English language than homophemous words, and so the mapping of phonemes to words will consist of far fewer options to choose from than the mapping of visemes to words and with the assistance of audio, one could decode phonemes with reasonable precision. However, in a setting where there is either poor audio, no audio and when only visual cues are utilised; there is still the problem of several phonemes sharing identical visemes to be overcome which in itself is a cause of confusion. 

When decoding visual speech in the form of phonemes or visemes,  a language model is required to subsequently determine which words were uttered along with a lexicon providing the choice of words. Various language models have been implemented for converting phonemes or visemes to words. This paper proposes a solution to the one-to-many mapping problem posed in automating lip reading using only visual by using perplexity analysis with a pre-trained autoregressive language model.

The Generative Pre-Training (GPT) Transformer \cite{b42} is an attention-transformer based language model designed to perform many tasks in Natural Language Processing, including sentence prediction and text classification.  It has been trained on a large dataset covering a wide range of vocabulary.

For the task of text classification, the GPT can calculate a metric called perplexity for any sentence in the form of a string of words to give an indicator of how grammatically sound the sentence is. A sentence that is more grammatically correct will generate a lower perplexity score because it is expected that combinations of words that are more grammatically correct would have a greater likelihood. The search for optimal perplexity is a technique that can applied to lip reading as a solution to viseme-to-word conversion, when a sequence of visemes can map to different combinations of words.  The most likely combination of words that correspond to a sequence of visemes will be that with the lowest perplexity score.

The other problem posed by lip reading using only visual cues is the segmentation of words because without audio, it is difficult to determine the boundary of a spoken word in terms of where it starts and stops and this problem is itself a focus of ongoing research. It is for this reason that two different scenarios have been investigated in this paper: one in which the boundaries of words are unknown and the other where word boundaries are unknown.

The rest of the paper is organised as follows: First in Section \ref{sec:Literature-Review}, a literature review is given describing the general approaches used in visual speech recognition, and the advantages and limitation of the different language models that can be used for lip reading systems that are not confined to fixed lexicons.  Then in Section \ref{sec:Methodology}, all the distinct components that make up the viseme-to-word conversion process are described including: the principle of perplexity analysis, the chunkification of visemes, the two aforementioned scenarios being modelled and the accuracy metrics used to evaluate the performance of the word detector. In Section \ref{sec:Experiment}, the classification results for the word detector are discussed and compared followed by concluding remarks given in Section \ref{sec:Conclusion} along with suggestions for further research.

\section{Literature Review}
\label{sec:Literature-Review}

The first automated lip reading systems to be constructed, had focused on classifying isolated speech segments in the form of digits and letters \cite{b13}\cite{b14}\cite{b15}\cite{b16}\cite{b17}, eventually moving on to longer speech segments in the form of words. The success of automated lip reading was constrained by the limited nature of training data available.  Initially, the only audio-visual datasets available were those with isolated speech segments, i.e., digits, alphabet and words\cite{b18}\cite{b19}\cite{b20}. Consequently, every speech segment was treated as a class.

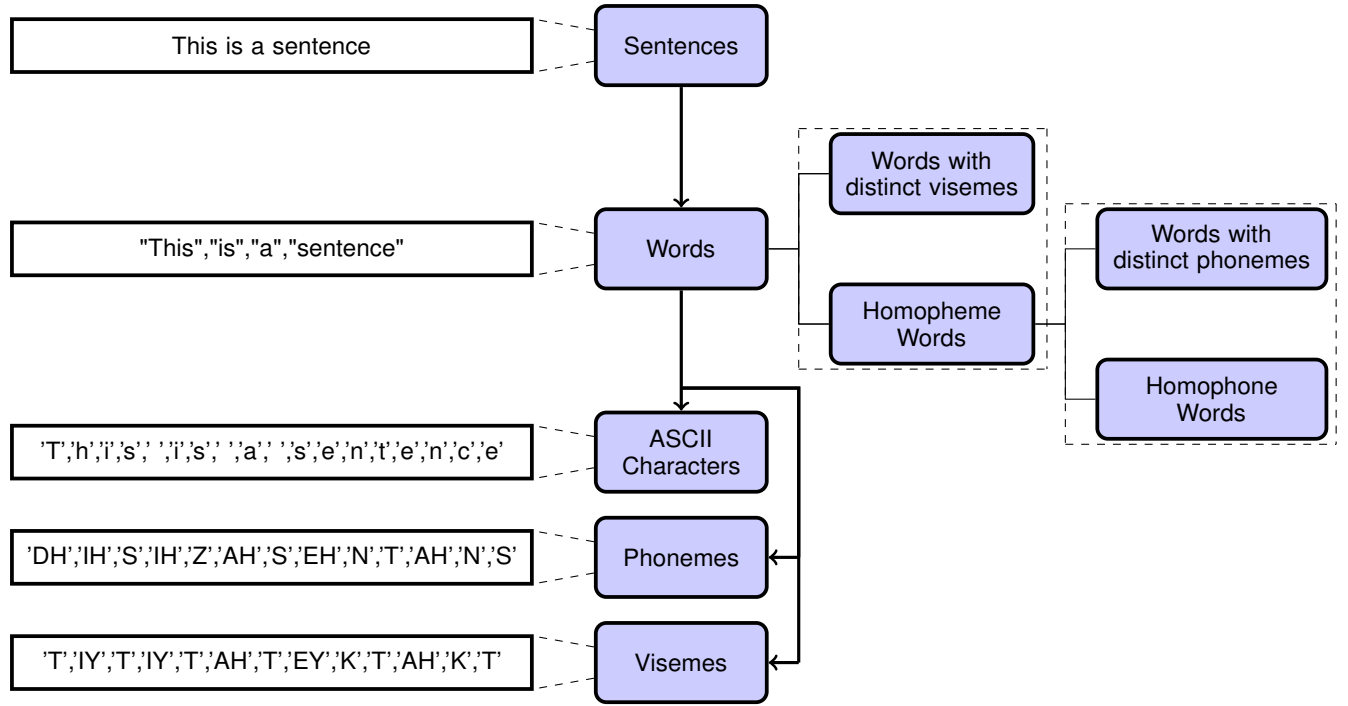
\begin{figure}[!htbp]
\begin{center}
\begin{tikzpicture}[node distance = 3.6cm, auto]
\tikzstyle{every node}=[font=\footnotesize\sffamily]
 % Place nodes
\node [block, rounded corners, minimum height=3.0em, text width=5.8em, line width=0.5mm] (Words) {Words};
\node [coordinate, right = 0.4cm of Words] (AS1){};
\node [coordinate, above = 1cm of AS1] (AS2){};
\node [coordinate, right = 0.4cm of AS2] (AS3){};
\node [coordinate, right = 0.4cm of Words] (BS1){};
\node [coordinate, below = 1cm of BS1] (BS2){};
\node [coordinate, right = 0.4cm of BS2] (BS3){};
\node [block, rounded corners, right = 0.4cm of AS2, minimum height=3.0em, text width=7em, line width=0.5mm] (Distinct Visemes) {Words with distinct visemes};
\node [block, rounded corners, right = 0.4cm of BS2, minimum height=3.0em, text width=7em, line width=0.5mm] (Homopheme Words) {Homopheme Words};
\node [block, rounded corners, below = 1.6cm of Words, minimum height=3.0em, text width=5.8em, line width=0.5mm] (ASCII Characters) {ASCII\\ Characters};
\node [block, rounded corners, above = 1.6cm of Words, minimum height=3.0em, text width=5.8em, line width=0.5mm] (Sentences) {Sentences};
\node [block, rounded corners, below = 3.0cm of Words, minimum height=3.0em, text width=5.8em, line width=0.5mm] (Phonemes) {Phonemes};
\node [block, rounded corners, below = 4.4cm of Words, minimum height=3.0em, text width=5.8em, line width=0.5mm] (Visemes) {Visemes};
\node [coordinate, below = 1.3cm of Words] (P2) {};
\node [coordinate, right = 0.4cm of Phonemes] (P3) {};
\node [coordinate, right = 0.4cm of Visemes] (P4) {};
\draw [->, very thick] (P3) -- (Phonemes);
\draw [->, very thick] (P4) -- (Visemes);
\draw [-, very thick] (P2) -| (P3);
\draw [-, very thick] (P2) -| (P4);

\node [block, rectangle, font=\footnotesize\sffamily, fill=white!20, left = 0.8cm of Sentences, minimum height=2.0em, text width=19.0em, line width=0.5mm] (Sentence example) {This is a sentence};
\node [coordinate, left = 0.8cm of Sentences, yshift=1em] (sentence-upper) {};
\node [coordinate, left = 0.8cm of Sentences, yshift=-1em] (sentence-lower) {};
\draw [dashed] (Sentences) -- (sentence-upper);
\draw [dashed] (Sentences) -- (sentence-lower);

\node [block, rectangle, font=\footnotesize\sffamily, fill=white!20, left = 0.8cm of Words, minimum height=2.0em, text width=19.0em, line width=0.5mm] (Words example) {"This","is","a","sentence"};
\node [coordinate, left = 0.8cm of Words, yshift=1em] (words-upper) {};
\node [coordinate, left = 0.8cm of Words, yshift=-1em] (words-lower) {};
\draw [dashed] (Words) -- (words-upper);
\draw [dashed] (Words) -- (words-lower);

\node [block, rectangle, font=\footnotesize\sffamily, fill=white!20, left = 0.8cm of ASCII Characters, minimum height=2.0em, text width=19.0em, line width=0.5mm] (Words example) {'T','h','i','s',' ','i','s',' ','a',' ','s','e','n','t','e','n','c','e'};
\node [coordinate, left = 0.8cm of ASCII Characters, yshift=1em] (ascii-upper) {};
\node [coordinate, left = 0.8cm of ASCII Characters, yshift=-1em] (ascii-lower) {};
\draw [dashed] (ASCII Characters) -- (ascii-upper);
\draw [dashed] (ASCII Characters) -- (ascii-lower);

\node [block, rectangle, font=\footnotesize\sffamily, fill=white!20, left = 0.8cm of Phonemes, minimum height=2.0em, text width=19.0em, line width=0.5mm] (Words example) {'DH','IH','S','IH','Z','AH','S','EH','N','T','AH','N','S'};
\node [coordinate, left = 0.8cm of Phonemes, yshift=1em] (phonemes-upper) {};
\node [coordinate, left = 0.8cm of Phonemes, yshift=-1em] (phonemes-lower) {};
\draw [dashed] (Phonemes) -- (phonemes-upper);
\draw [dashed] (Phonemes) -- (phonemes-lower);

\node [block, rectangle, font=\footnotesize\sffamily, fill=white!20, left = 0.8cm of Visemes, minimum height=2.0em, text width=19.0em, line width=0.5mm] (Words example) {'T','IY','T','IY','T','AH','T','EY','K','T','AH','K','T'};
\node [coordinate, left = 0.8cm of Visemes, yshift=1em] (visemes-upper) {};
\node [coordinate, left = 0.8cm of Visemes, yshift=-1em] (visemes-lower) {};
\draw [dashed] (Visemes) -- (visemes-upper);
\draw [dashed] (Visemes) -- (visemes-lower);

\node [coordinate, right = 0.6cm of ASCII Characters] (AS4){};
\node [coordinate, above = 0.6cm of AS4] (AS5){};
\node [coordinate, right = 0.6cm of AS5] (AS6){};
\node [coordinate, right = 0.6cm of ASCII Characters] (BS4){};
\node [coordinate, below = 0.6cm of BS4] (BS5){};
\node [coordinate, right = 0.6cm of BS5] (BS6){};

\node [coordinate, above = 1.6cm of AS1] (AT1){};
\node [coordinate, below = 1.6cm of AS1] (BT1){};   
\node [coordinate, right = 3.3cm of AT1] (AT2){};
\node [coordinate, right = 3.3cm of BT1] (BT2){};
\node [coordinate, above = 1.0cm of AS4] (AT4){};
\node [coordinate, below = 1.0cm of AS4] (BT4){};
\node [coordinate, right = 4.0cm of AT4] (AT5){};
\node [coordinate, right = 4.0cm of BT4] (BT5){};
\node [coordinate, left = 0.0cm of AT4] (AT0){};
\node [coordinate, left = 0.0cm of BT4] (BT0){};
\node [coordinate, below = 3.4cm of Words] (BT6){};
\node [coordinate, right = 1.4cm of BT6] (BT7){};
\node [coordinate, left = 1.4cm of BT6] (BT8){};
\node [coordinate, above = 8.0cm of BT7] (AT7){};
\node [coordinate, above = 8.0cm of BT8] (AT8){};

\node [coordinate, right = 0.4cm of Homopheme Words] (AU0){};
\node [coordinate, right = 0.4cm of Homopheme Words] (BU0){};

\node [coordinate, right = 0.4cm of Homopheme Words] (AU1){};
\node [coordinate, above = 1cm of AU1] (AU2){};
\node [coordinate, right = 0.4cm of AU2] (AU3){};
\node [coordinate, right = 0.4cm of Homopheme Words] (BU1){};
\node [coordinate, below = 1cm of BU1] (BU2){};
\node [coordinate, right = 0.4cm of BU2] (BU3){};
\node [coordinate, above = 1.6cm of AU1] (AV1){};
\node [coordinate, below = 1.6cm of AU1] (BV1){};   
\node [coordinate, right = 3.6cm of AV1] (AV2){};
\node [coordinate, right = 3.6cm of BV1] (BV2){};

    % Draw edges
    \draw [-] (Words) -- (AS1);
    \draw [-] (Words) -- (BS1);
    \draw [-] (AS1) |- (AS3);
    \draw [-] (BS1) |- (BS3);
    \draw [->, very thick] (Words) -- (ASCII Characters);
    \draw [<-, very thick] (Words) -- (Sentences);
    \draw [dashed] (AS1) -- (AT1);
    \draw [dashed] (AS1) -- (BT1);
    \draw [dashed] (AT1) -- (AT2);
    \draw [dashed] (BT1) -- (BT2);
    \draw [dashed] (AT2) -- (BT2);

    %\draw [dashed] (BT7) -- (BT8);
    %\draw [dashed] (AT7) -- (AT8);
    %\draw [dashed] (AT7) -- (BT7);
    %\draw [dashed] (AT8) -- (BT8);
    
    \draw [-] (AS1) |- (AS3);
    \draw [-] (BS1) |- (BS3);
    
    \draw [-] (Homopheme Words) -- (AU0);
    \draw [-] (Homopheme Words) -- (BU0);
    
    \draw [dashed] (AU1) -- (AV1);
    \draw [dashed] (AU1) -- (BV1);
    \draw [dashed] (AV1) -- (AV2);
    \draw [dashed] (BV1) -- (BV2);
    \draw [dashed] (AV2) -- (BV2);

    \node [block, rounded corners, right = 0.4cm of AU2, minimum height=3.0em, text width=8em, line width=0.5mm] (Distinct Visemes) {Words with distinct phonemes};
    \node [block, rounded corners, right = 0.4cm of BU2, minimum height=3.0em, text width=8em, line width=0.5mm] (Distinct Visemes) {Homophone\\ Words};
    \draw [-] (AU1) |- (AU3);
    \draw [-] (BU1) |- (BU3);

\end{tikzpicture}
\end{center}
\vspace{-0.8em}
\caption{Hierarchy of speech classification.}
\label{fig:hierarchy}
\end{figure}

More recent datasets not only consist of speech segments in continuous speech, but they also consist of entire sentences that cover an extensively vocabulary range \cite{b6}. One could potentially assign a class to every possible word, which would help to distinguish between words that look identical because there is more temporal information available compared with shorter speech segments such as digits letters; but to do so would not be feasible due the computational bottleneck created from the existence of too many classes. It is for this reason that sentence-based lip reading systems use ASCII characters as classes whereby sentences are decoded as sequences of ASCII characters. Figure \ref{fig:hierarchy} gives a hierarchy of the different classification schema that can be used for lip reading.

The use of ASCII characters as a class system in speech recognition relies on the conditional dependence relationship that exists between characters. However, ACII characters are not always phonetic due to the presence of silent letters and digraphs meaning that they do not have a direct correlation with visual speech. The main limitation in using either words or ASCII characters as classes is that a speech recognition system would only be able to words that have appeared in the training phase because for word classification, the word needs to be encoded as a class and be present in the training phase;  and for ASCII classification, the prediction of words is based on combinations of characters having been presented in the training phase as patterns.

A viseme is a unit of visual speech and the visual equivalent of a phoneme \cite{b10}\cite{b11} which is a spoken unit of speech that can be represented by an acoustic signal. According to Hazen \cite{b12}, there are roughly 40 phonemes in the English language with only around a dozen distinguishable visemes. This means that words that look the same when spoken, i.e. homophemes or homovisemes are far more frequent than homophone words, i.e. those that sound the same when spoken. Roughly half of words in the English have distinct visemes while the other half are homopheme words.

Visemes as a classification schema have many advantages to words or ASCII characters in that they require fewer classes, do not need pre-trained lexicons and can be generalised to other languages. There are however limitations to using visemes in that not only does the one-to-many mapping problem mean that different words share identical visemes, but the shorter duration of visemes in comparison to words means that less temporal information is available to distinguish between classes.
There are different sequential routes (summarized below) that can be utilised for lip reading systems whereby one has to determine precisely which words were spoken given the visual feature input from a person’s moving lips. The first route is a lip reading system that outputs words directly and will generally use a classification schema were each word is treated as a class, or a sequence of ASCII characters:
\vspace{-0.8em}
\begin{enumerate}
\item Visual Features \verb|->| Words
\vspace{-0.4em}
\item Visual Features \verb|->| Phonemes \verb|->| Words
\vspace{-0.4em}
\item Visual Features \verb|->| Visemes \verb|->| Words
\end{enumerate}

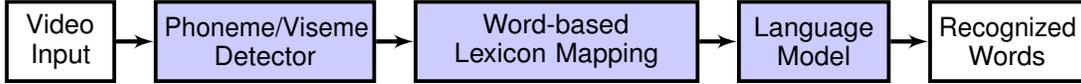
\begin{figure}[!htbp]
\begin{center}
\begin{tikzpicture}[node distance = 1.6cm, auto]
\tikzset{every node}=[font=\sffamily]

%\tikzset{every node}=[font=\itshape]
% Place nodes
	\node [block, minimum height=3.0em, text width=7.6em, line width=0.5mm, fill=blue!20] (Step1) {Phoneme/Viseme Detector};
	\node [coordinate, left = 1.1cm of Step1] (C1) {};
	\node [block, minimum height=3.0em, text width=3.4em, line width=0.5mm, fill=white!20, left = 0.5cm of Step1] (Step0) {Video \\Input};
 	\node [block, minimum height=3.0em, text width=10.0em, line width=0.5mm, fill=blue!20, right = 0.5cm of Step1] (Step2) {Word-based\\ Lexicon Mapping};
	\node [block, minimum height=3.0em, text width=5.0em, line width=0.5mm, fill=blue!20, right = 0.5cm of Step2] (Step3) {Language\\ Model};
	\node [block, minimum height=3.0em, text width=5.2em, line width=0.5mm, fill=white!20, right = 0.5cm of Step3] (Step4) {Recognized Words};
	\path [line, ultra thick] (Step0) -- (Step1);
	\path [line, ultra thick] (Step1) -- (Step2);
	\path [line, ultra thick] (Step2) -- (Step3);
	\path [line, ultra thick] (Step3) -- (Step4);

\end{tikzpicture}
\end{center}
\vspace{-0.8em}
\caption{Framework for Lip Reading Systems which classify phonemes or visemes.}
\label{fig:framework}
\end{figure}

There are a vast number of automated lip reading systems devoted to the first category mentioned above. However, such approaches are often confined to a working set of vocabulary and only predict words that have appeared in the training phase. The second and third category will predict phonemes and visemes in sequence and then apply a language model, which is a probability distribution over sequences of words, to determine the spoken set of words. Lip Reading systems that fall into the second and third category generally follow a procedure illustrated in Figure \ref{fig:framework}.

The second category of lip reading systems is useful for when there is some sound available and because there are a limited number of acoustic sounds or phonemes that exist, one can generalise the system to be able to decode words that do not fall into a set list of vocabulary and that have not appeared in the training phase. This however is only the case for when sound is available because in the absence of audio, several phonemes can look identical on a speaker.

In the absence of sound, one could only distinguish homopheme words when there is contextual information available. Homopheme words can be further broken down into homophone words and words with distinct phonemes (Figure \ref{fig:hierarchy}). The language model used in this paper falls into the third category and, visemes for the input of the language model which is viseme-to-word detector using perplexity analysis to determine which set of words were uttered.

The problem of one-to-many mapping is addressed in several papers that have predicted words from only visual cues where speech is decoded either without the assistance of audio or with a corrupted audio. Words can be decomposed into either phonemes or visemes. 

When there is no sound available, it is possible to classify visemes and the mapping of visual lip movements to visemes is one-to-one and direct. One can still attempt to predict phonemes in the absence of sound, however this requires analysis of phonemes in combination to deduce the uttered phonemes based on conditional probability and this requires context.

Whilst viseme classification has a direct mapping and uses fewer classes than phonemes, there are advantages of performing phoneme classification, one of them being that the class imbalance is smaller. Converting from phonemes to words is also advantageous to the conversion of visemes to words because there are fewer homophone words than homopheme words which means that phoneme conversion has more discriminative power than viseme conversion.

There are two types of language model: statistical language models and neural language models. Statistical language models predict words based on the preceding words in the sequence based on the Markov assumption. The language model used in this paper is a neural language model and is provided by a pre-trained attention-based transformer.

The conversion of visemes or phonemes to words will performed using a trained language model to predict the most likely words that have been uttered given the identity of visemes or phonemes. Weighted Finite State Transducers(WFSTs)\cite{b21} and Hidden Markov Models(HMMs)\cite{b22} are some examples of algorithms used to implement language models based on Markov chains, which assume that each word in a sentence depends only its predecessors.

An N-gram language model is a language model that predicts sequences of words according to the Markov process where the probability of the next word in a sequence is predicted based on the previous $N-1$ words. Eq. \ref{eq1} gives the ideal chain rule of probability $P$ to apply to any language model with a sequence of $K$ words. However as $K$ increases, the computation because impossible so statistical language models use the Markov assumption given in Eq. \ref{eq2}.

\begin{equation} P(w_1,w_2,...,w_K)=\prod_{i} P(w_i |w_1,w_2,...,w_{i-1} ) \label{eq1}
\end{equation}
\begin{equation} P(w_1,w_2,...,w_K)=\prod_{i} P(w_i |w_{i-N+1},...,w_{i-1} ) \label{eq2}
\end{equation}

Miao\cite{b23} and Assael\cite{b24} both use WFSTs to implement 3-gram and 5-gram language models respectively for the conversion of phonemes to words. Cabellero\cite{b25} and Howell\cite{b26} both use a WFST to implement a bigram language models when converting phonemes to words and the phoneme classification was performed for dysarthric speech. Dysarthria is a condition where there is a poor articulation of phonemes and as a consequence, lots of phonemes that share identical lip movements are confused, making the task of lip reading similar to situation of only being able to classify the uttered visemes\cite{b27}. However, both Cabellero\cite{b25} and Howell\cite{b26}  do use confusion modelling to correct for possible misclassified phonemes by detecting possible deletion, insertions and substitutions that could have taken place between the ground truth to the predicted result. Thangthai and Harvey\cite{b28} classified visemes as used a HMM to implement a bigram language model to convert visemes to words.

N-gram language models are limited in comparison to neural language models because they need a large value of $N$ to produce an accurate language model which requires lots of computational overhead. N-grams are also a sparse representation of language because they are based on the probability of words in combination, and so would naturally give a zero probability to combinations of words that have not previously appeared. Furthermore N-grams fail to accurately predict semantic and syntactic details of sentences.

Howell at el.\cite{b26} do tackle the problem of data sparsity by adding Katz smoothing to deal with the problem of unseen bigrams whereby unseen bigrams are approximated. However a bigram language model will still fail to predict unseen trigrams and an N-gram model will fail to predict an unseen (N+1)-gram. Though one in theory can just increase the value of $N$,  but this would lead to an exponential explosion in the computation required.
 
There is no official standard convention for defining precise visemes or even the precise total number of visemes and different approaches to viseme classification have used varying numbers of visemes as part of their conventions with different phoneme-to-viseme mappings \cite{b30}\cite{b31}\cite{b32}\cite{b33}\cite{b34}\cite{b35}.  All the different conventions consist of consonant visemes, vowel visemes and one silent viseme but Lee and Yook’s \cite{b30} mapping convention appears to be the most favoured for speech classification and it is the one that has been utilised for this paper. It is however accepted that there are multiple phonemes that are visually identical on any given speaker \cite{b36}\cite{b37}.

To automate lip reading in real-time, it is necessary to predict spoken words in continuous speech and when only visual information is present, it is difficult to determine where each word starts and stops. There is an extensive literature discussing methods that can be used to segment words and locate word boundaries using only visual clues and these include prosodic cues such as stress or intonation patterns such as falling pitches at word boundaries \cite{b38}. The details are beyond the scope of this paper; however, interested readers can refer to, for example, Mitchel and Weiss’s paper \cite{b39} for more details.  

The recognition of when a word starts and stops is another challenge posed in visual speech recognition and for this reason, two scenarios have been modelled in this paper: one where the word beginnings and endings are known and the other where the beginnings and endings are unknown. 

\section{Methodology}
\label{sec:Methodology}

Because of the ongoing attempts to segment visual speech by words to know where they start and stop without the assistance the assistance of audio, two different scenarios have been modelled in this paper. In one scenario, it assumed that the word boundaries are known where the position of word boundaries is utilised, while in the second scenario, it is assumed that the word boundaries are not known. Therefore, it is expected that more sentences will be predicted in Scenario 2. Modelling these two scenarios is important because while the former results in fewer sentences to choose between, the determination of word boundaries still remains an ongoing challenge, so it is still a necessity to consider the second scenario.

\subsection{Pre-processing}
\label{subsec:Pre-processing}

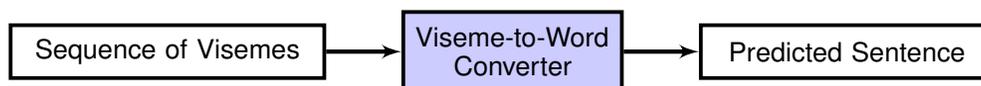
\begin{figure}[!htbp]
\begin{center}
\begin{tikzpicture}[node distance = 1.6cm, auto]
\tikzset{every node}=[font=\sffamily]

%\tikzset{every node}=[font=\itshape]
% Place nodes
	\node [block, minimum height=3.0em, text width=7.6em, line width=0.5mm, fill=blue!20] (Step1) {Viseme-to-Word Converter};
	\node [coordinate, left = 1.1cm of Step1] (C1) {};
	\node [block, minimum height=2.0em, text width=11.2em, line width=0.5mm, fill=white!20, left = 1.0cm of Step1] (Step0) {Sequence of Visemes};
	\node [block, minimum height=2.0em, text width=10.4em, line width=0.5mm, fill=white!20, right = 1.0cm of Step1] (Step4) {Predicted Sentence};
	\path [line, ultra thick] (Step0) -- (Step1);
	\path [line, ultra thick] (Step1) -- (Step4);

\end{tikzpicture}
\end{center}
\vspace{-1.0em}
\caption{The input and output of viseme-to-word converter.}
\label{fig::in-out}
\end{figure}

The objective of the experimentation described in this paper is to determine what words have been spoken where only the visemes uttered by the individual are known, The Viseme-to-Word converter takes a sequence of visemes and input and gives a predicted sentence as the output as presented in Figure \ref{fig::in-out}.

To experimentally examine this procedure, sentences from lip reading datasets are converted to sequences of visemes. Possible selections of words are then predicted from those visemes and just like a standard machine learning algorithm, where there are features and labels with feature inputs and predicted outputs; visemes make up the feature input and predicted words make up the predicted outputs.

\begin{figure}[!htbp]
\begin{center}
\includegraphics[scale = 0.56]{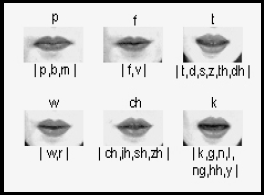}
\includegraphics[scale = 0.56]{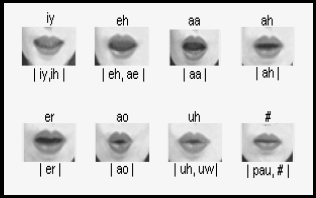}
\vspace{-0.8em}
\caption{The six consonant visemes on the \textbf{left} and the 7 vowel visemes and silent viseme on the \textbf{right}. \cite{b30}}
\label{ref::Figure-Viseme}
\end{center}
\end{figure}

Table \ref{table:Table-Viseme} and Figure \ref{ref::Figure-Viseme} highlight the one-to-many mapping problem that exists between visual lip movements and acoustic sounds. There are a set of distinct lip movements called visemes, and each viseme can map to multiple sounds or phonemes. Two pre-processing stages are required to create the feature inputs from lip reading sentence data:
\vspace{-0.8em}
\begin{itemize}
\item Step 1: Mapping of words to phonemes using the Carnegie Mellon Pronouncing Dictionary \cite{b41}
\vspace{-0.4em}
\item Step 2: Mapping the identified phonemes to visemes according to Lee and Yook’s approach \cite{b30}
\end{itemize}

\begin{figure}[!htbp]
\begin{center}
\begin{tikzpicture}[node distance = 2cm, auto]
\tikzstyle{every node}=[font=\footnotesize\sffamily]
 % Place nodes
 \node [block3, fill=white!20, text width=19.0em] (Word1) {What time is it?};
 \node [empty3, text width=12.0em, left = 0.6cm of Word1] (Stage1) {Words};
 \node [block3, fill=white!20, text width=19.0em, below = 0.3cm of Word1] (Word2) {{['w', 'ah', 't'], ['t', 'ay', 'm'], ['ih', 'z'], ['ih', 't']}};
  \node [empty3, text width=12.0em, left = 0.6cm of Word2] (Stage2) {Phonemes};
 \node [block3, fill=white!20, text width=19.0em, below = 0.3cm of Word2] (Word3) {{['w', ‘ah', 't'], ['t', 'ah', 'p'], ['iy', 't'], ['iy', 'y']}};
 \node [empty3, text width=12.0em, left = 0.6cm of Word3] (Stage2) {Visemes};
 \node [block3, fill=white!20, text width=19.0em, below = 0.3cm of Word3] (Word4) {{'w', 'ah', 't', 't', 'ah', 'p', 'iy', 't', 'iy', 't'}};
  \node [empty3, text width=12.0em, left = 0.6cm of Word4] (Stage2) {Concatenated Visemes};
 \path [line] (Word1) -- (Word2);
 \path [line] (Word2) -- (Word3);
 \path [line] (Word3) -- (Word4);
\end{tikzpicture}
\end{center}
\vspace{-2.0em}
\caption{Process map of the conversion of words to visemes.}
\label{fig:viseme-phoneme}
\end{figure}
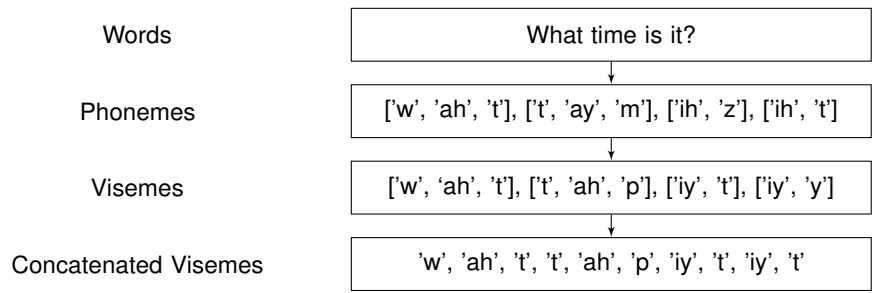

An example of a viseme conversion process covering the two aforementioned steps is given in Figure \ref{fig:viseme-phoneme} with the sentence example "what time is it?". For Scenario 2, the final step is not needed.

\begin{table}[!htbp]
\begin{center}
\caption{Viseme-to-Phoneme Mappings.}
\vspace{-0.6em}
\begin{tabular}{|c|c|c|}
\hline
\textbf{Viseme Class} & \textbf{Viseme Type} & \textbf{Phonemes Set}\\ \hline
p & consonant & b, p, m\\ \hline
t & consonant & d, t, s, z, th, dh\\ \hline
k & consonant & g, k, n, ng, l, y, hh\\ \hline
ch & consonant & jh, ch, sh, zh\\ \hline
f & consonant & f, v\\ \hline
w & consonant & r, w\\ \hline
iy & vowel & iy, ih\\ \hline
ey & vowel & eh, ey, ae\\ \hline
aa & vowel & aa, aw, ay, ah\\ \hline
ah & vowel & ah\\ \hline
ao & vowel & ao, oy, ow\\ \hline
uh & vowel & uh, uw\\ \hline
er & vowel & er\\ \hline
s & silent character & sil\\ \hline
\end{tabular}
\label{table:Table-Viseme}
\end{center}
\end{table}

\vspace{-0.0em}

Once inputs have been pre-processed, the word detection is implemented on viseme sequences. The concatenation of visemes is only performed for Scenario 2 whereby the visemes undergo chunkification (explained in Subsection \ref{subsec:Chunkification}) to output all possible viseme clusters that could have been uttered. Spoken sentences are predicted from uttered visemes using perplexity calculation performed using an iterative process described in Subsection \ref{subsec:Iterations}.

%\begin{equation}
%	\xi _{ij}(t)=P(x_{t}=i,x_{t+1}=j|y,v,w;\theta)= {\frac {\alpha _{i}(t)a^{w_t}_{ij}\beta _{j}(t+1)b^{v_{t+1}}_{j}(y_{t+1})}{\sum _{i=1}^{N} \sum _{j=1}^{N} \alpha _{i}(t)a^{w_t}_{ij}\beta _{j}(t+1)b^{v_{t+1}}_{j}(y_{t+1})}}
%\end{equation}

%\paragraph{Paragraph}
%\lipsum[7]

\subsection{Word Detector}
\label{subsec:Word-Detector}

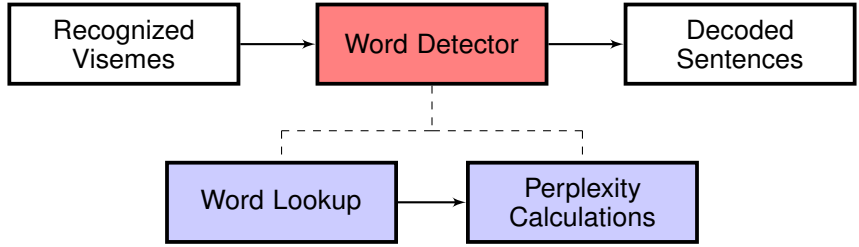
\begin{figure}[!htbp]
\begin{center}
\begin{tikzpicture}[node distance = 1.6cm, auto]
\tikzset{every node}=[font=\sffamily]

%\tikzset{every node}=[font=\itshape]
% Place nodes
	\node [block, minimum height=3.0em, line width=0.5mm, fill=white!50] (Step4) {Recognized Visemes};
	\node [block, minimum height=3.0em, line width=0.5mm, fill=red!50, right = 1.0cm of Step4] (Step5) {Word Detector};
	\node [block, minimum height=3.0em, line width=0.5mm, fill=white!20, right = 1.0cm of Step5] (Step6) {Decoded Sentences};
	\path [line, thick] (Step4) -- (Step5);
	\path [line, thick] (Step5) -- (Step6);
	
	\node [coordinate, below = 0.6cm of Step5] (C0){};
	\node [coordinate, left = 2.0cm of C0] (CL){};
	\node [coordinate, right = 2.0cm of C0] (CR){};
	\draw [dashed] (Step5) -- (C0);
	\draw [dashed] (C0) -- (CL);
	\draw [dashed] (C0) -- (CR);
	\node [block, minimum height=3.0em, line width=0.5mm, fill=blue!20, below = 0.4cm of CL] (WL) {Word Lookup};
	\node [block, minimum height=3.0em, line width=0.5mm, fill=blue!20, below = 0.4cm of CR] (PCC) {Perplexity Calculations};
	\draw [dashed] (CL) -- (WL);
	\draw [dashed] (CR) -- (PCC);
	\path [line, thick] (WL) -- (PCC);

\end{tikzpicture}
\end{center}
\vspace{-0.8em}
\caption{The input and output of viseme-to-word converter.}
\label{fig::word-detector}
\end{figure}

Every word in a sentence contains a set of visemes and therefore can be mapped to a cluster of visemes, such that a cluster of visemes is a set of visemes which make up a word. Because a cluster of visemes can map to several different words, the combination of the words that were uttered by the speaker still needs to be deciphered and to do this requires a viseme-to-word conversion process. The solution to the one-to-many mapping problem in the viseme-to-word conversion process is to select the most likely combination of words. The general procedure for converting visemes to words with a series of stages is given in Figure \ref{fig::word-detector}.

The first stage of the Word Detection is the World Lookup stage. Every single cluster of visemes needs to be mapped to a set of words containing those visemes according to the mapping given by the Carnegie Mellon Pronouncing (CMU) Dictionary.

Once the word lookup stage is performed, the next stage of Word Detection is the Perplexity Calculations. The numerous possible choices of words that map to the visemes are combined, and perplexity iterations are performed to determine which combination of words is most likely to correspond to the uttered sentence, given the visemes recognised. Naturally, the sentence that is most grammatically correct will have the highest likelihood and perplexity is one metric that can be used to compare sentences to determine which is most grammatically sound. When performing a conversions of visemes to words, the following selection rules are implemented:
\vspace{-0.8em}
\begin{enumerate}
\item If a viseme sequence has only 1 cluster matching to one word, that one word is selected as the output.
\vspace{-0.4em}
\item If a viseme sequence has only 1 cluster matching to several words, that word with largest expectation is selected as the output. This is determined by word rankings found in the Corpus of Contemporary American English(COCA)\cite{b29}.
\vspace{-0.4em}
\item If a viseme sequence has more than 1 cluster, the words matching to the first two clusters are combined in every possible combination for the first iteration
\begin{enumerate}
\item The combinations with the lowest 50 perplexity scores are kept.
\item These combinations are in turn combined with the words matching to the next viseme cluster.
\item The combinations with the lowest 50 perplexity scores are kept and the iterations continue for the remaining clusters of the sequence until the end of the sequence is reached.
\end{enumerate}
\end{enumerate}

The selection of the lowest 50 perplexity scores at each iteration is based on an implementation of a local beam search with width 50. In practice, it would be computationally expensive to perform an exhaustive search. Therefore, a beam search has been implemented to reduce the computational overhead, and the beam width itself is an arbitrary figure chosen as a compromise between accuracy and computational efficiency.

Eqs. \ref{eq3} to \ref{eq6} describe the probabilistic relationship between the observed visemes and the words spoken; where $V$ is the spoken sequence of viseme clusters, $v_i$ corresponds to every ith cluster, $W_C$ represents any given combination of words and $w_i$ corresponds to every ith word within the string of words. The string of words $\check{W}$ that is to be selected will be the combination that has the maximum likelihood given the identity of the viseme clusters for every combination $C$ that falls within the set of combinations $C^*$. The sequence of visemes clusters given in Eq. \ref{eq3} maps to any possible combination of words as given in Eq. \ref{eq4}, and the solution to predicting the sentence spoken is the combination of words given the recognised visemes which has the greatest probability as expressed in Eqs. \ref{eq5} and \ref{eq6}.

\begin{equation} V=(v_1,v_2,...,v_N)=\sum_{i=1}^{N}v_i \label{eq3}
\end{equation}
\begin{equation} W_C=(w_1,w_2,...,w_N)_C=\sum_{i=1}^{N}w_i                                                                                    
\label{eq4}
\end{equation}
\begin{equation} \check{W}==\arg\max_{C\epsilon C^*}\left[P(W|V)\right]_{C}  
\label{eq5}
\end{equation}
\begin{equation} \check{w}_1,\check{w}_2,...,\check{w}_N=\arg\max_{C\epsilon C^*}[P(w_1,w_2,...,w_N |v_1,v_2,...,v_N )]_C 
\label{eq6}
\end{equation}

If the identity of observed visemes is known, the probability of the viseme sequence in Eq. \ref{eq3}  should equal to being equal to 1 (Eq. \ref{eq7}) and the choice of words predicted according to Eq. \ref{eq6} gets reduced to the expression given in Eq. \ref{eq8}. 

\begin{equation} P(v_1,v_2,…,v_N )=1 \label{eq7}\end{equation}
\begin{equation}\check{w}_1,\check{w}_2,...,\check{w}_N=\arg\max_{C\epsilon C^*}[P(w_1,w_2,...,w_N)]_C  \label{eq8}\end{equation}

\subsection{Perplexity}
\label{subsec:Perplexity}

Eqs. \ref{eq9} to \ref{eq12} below describe the relationship between the perplexity $PP$, entropy $H$ and probability $P(w_1,w_2,…,w_N)$ of a particular sequence of $N$ words $(w_1,w_2,…,w_N)$. The word detector uses the GPT to calculate PP expressed as the exponentiation of $H$ in Eq. \ref{eq9}. The per-word entropy $H$ is related to the probability $P(w_1,w_2,…,w_N)$ of words $(w_1,w_2,…,w_N)$ belonging to a vocabulary set $W$, and is calculated as a summation is over all possible sequences of words. If the source is ergodic, the expression for $\hat{H}$ in Eq. \ref{eq10} gets reduced to that in Eq. 11. The value of $P(w_1,w_2,…,w_N)$ resulting in the choice of words selected as the output for Eq. \ref{eq8} also results in the minimisation of entropy in Eq. \ref{eq11}, further resulting in the minimisation of perplexity given in Eq. \ref{eq12}.

\begin{equation} PP=e^H \label{eq9}\end{equation}
\begin{equation} \hat{H}=-\lim_{N \to \infty} \frac{1}{N} \sum_{w_1,w_2,...w_N}P(w_1,w_2,...,w_N)\ln{P(w_1,w_2,...,w_N)} \label{eq10}\end{equation}
\begin{equation} \hat{H}=-\frac{1}{N}\ln{P(w_1,w_2,...,w_N)} \label{eq11}\end{equation}
\begin{equation} PP=P(w_1,w_2,...,w_N)^{-\frac{1}{N}} \label{eq12}\end{equation}

A language model, which is a probability distribution over sequences of words, can be measured on the basis of the entropy of its output from the field of information theory \cite{b43}. Perplexity is a measure of the quality of a language model, because a good language model will generate sequences of words with a larger probability of occurrence resulting in a smaller perplexity.

The Transformer model used for the word detector is the pre-trained Generative Pre-Training (GPT) Transformer [42] – a multi-layer decoder and a variant of the transformer used in \cite{b40}. It consists of repeated blocks of multi-headed self-attention followed by position-wise feedforward layers. The architecture is typically used for sentence prediction; however, the architecture itself here is not used for direction classification, rather its purpose is for perplexity calculations that are required for word selection where visemes are converted to words. Visemes from the previous step are sequentially matched to words and the most probable sentence is chosen according to that with the minimum perplexity score. The perplexity score is calculated by taking the exponentiation of the cross-entropy loss when the GPT is evaluated on a sentence, a beam width of 50 is used.

\subsection{Chunkification}
\label{subsec:Chunkification}

For Scenario 2, it is assumed that the spaces between words are unknown and so all the visemes will be joined together in long sequences. Viseme sequences will then be segmented into possible clusters of visemes such that each viseme cluster maps to at least one word that can be found in the CMU dictionary. The segmentation process is performed using recursion and the viseme-to-word conversion process is performed for all possible sequences of visemes. For  Scenario 1, viseme-to-word conversion is only performed for clusters in the original segmented format corresponding to the actual phrases, so no chunkification is performed.

The recursion process is given in Listing \ref{listing::find_possible_chunks} in a function called \textbf{"find possible chunks"}. For every viseme sequence, a chunk of visemes is taken off of the beginning of the visemes sequence and checked to see if it matches any of the words found in the CMU dictionary\cite{b41} (this is the process covered by the function \textbf{"find shortest prefix"} in Listing \ref{listing::find_shortest_prefix}). If a match is found, it gets taken off the sequence and the \textbf{find shortest prefix} process is repeated for the -remaining sequence. The same process is carried for the remaining sequence in recursive manner with the remaining sequence getting shorter every time a chunk of visemes is matched to a word. If no match is found for the initial chunk, the length of the chunk is incremented until a match is found. If no match can be found while the length of the last chunk continues to be incremented, the loop is broken and the length of the previous chunk is incremented.

\begin{figure}[!htbp]
\begin{center}
\includegraphics[scale = 1.0]{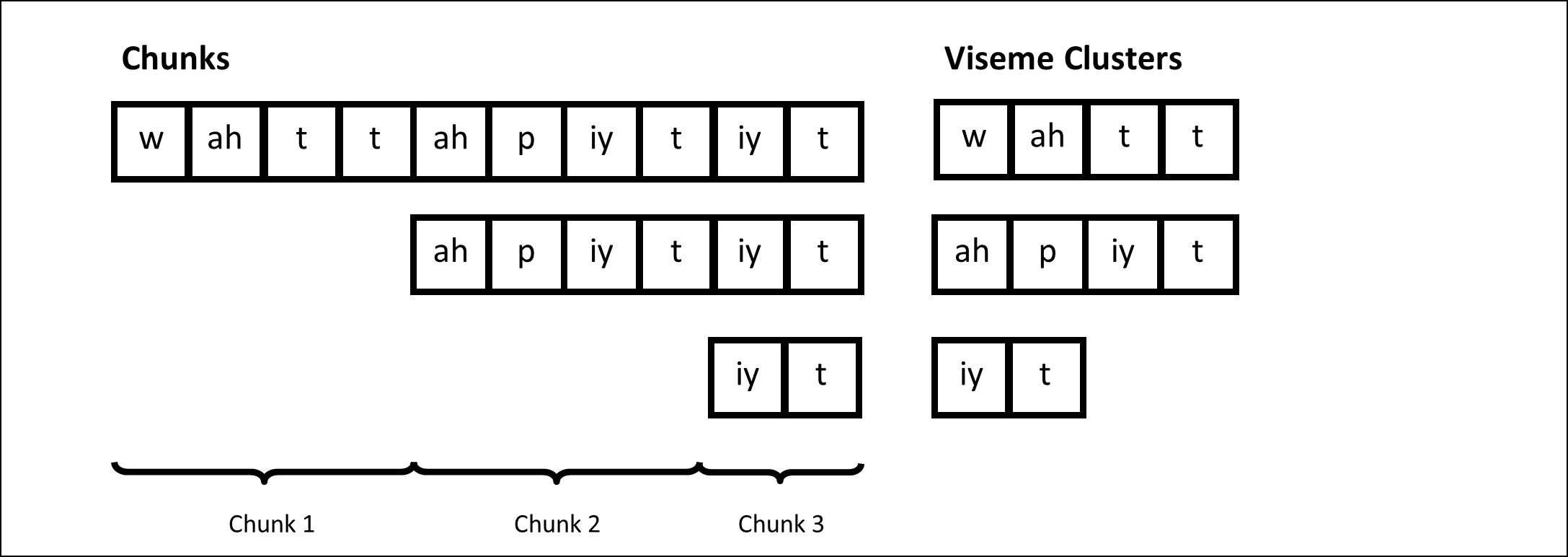}
\vspace{-1.2em}
\caption{Process map of the chunkification of a viseme sequence.}
\label{fig::Chunk1}
\end{center}
\end{figure}

\begin{figure}[!htbp]
\begin{center}
\includegraphics[scale = 1.0]{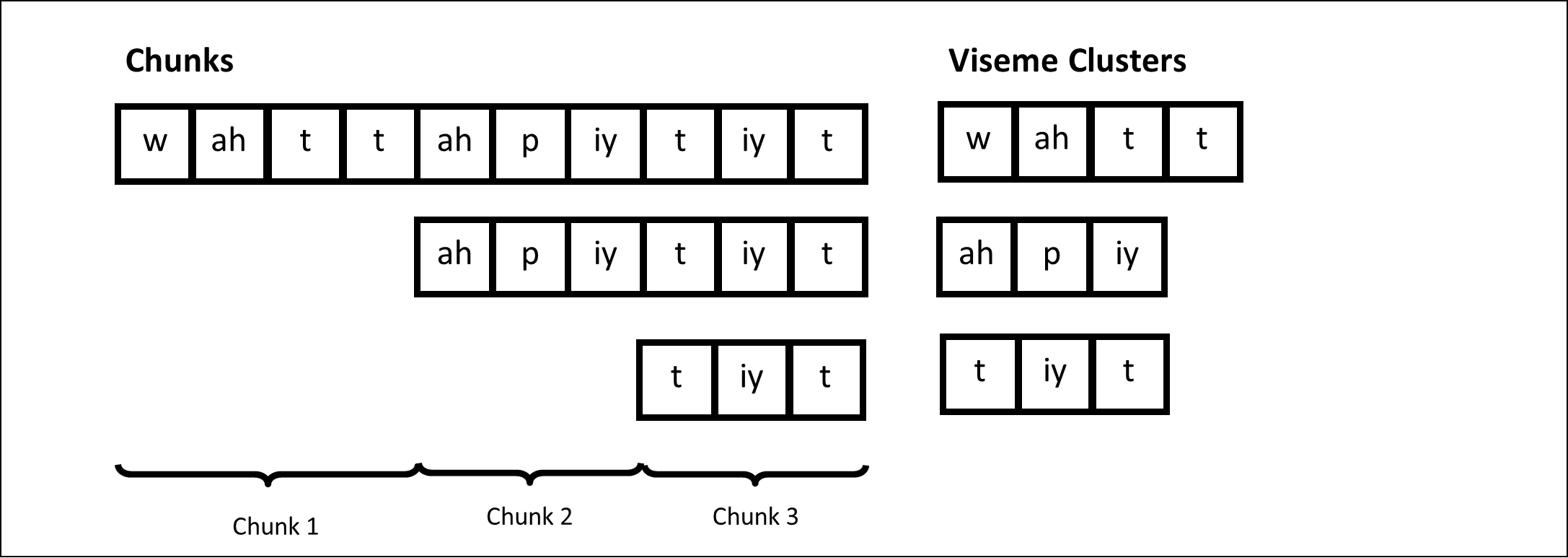}
\vspace{-1.2em}
\caption{Process map of the chunkification of the next viseme sequence.}
\label{fig::Chunk2}
\end{center}
\end{figure}

If clusters can be extracted for the whole viseme sequence, the sequence of clusters gets added to a list of cluster sequences and the chunkification continues until no more matches can be found. Figure \ref{fig::Chunk1} gives an example of viseme clusters that make up the sentence "what time is it", with the words corresponding to those clusters listed in Table \ref{table:Word-matches-1}. Figure \ref{fig::Chunk2} gives an example of the set of clusters that would be extracted at the next iteration of the chunkification process whereby the length second last cluster has increased by 1 viseme. Table \ref{table:Word-matches-2} shows the corresponding word matching for those clusters.

\begin{lstlisting}[language=Python, caption={Find Possible Chunks}\label{listing::find_possible_chunks}]
def find_possible_chunks(visemes, current=[]):
    successes = []
    n = 0
    while n < len(visemes):
        n = find_shortest_prefix(visemes, n)

        # no prefix looks like a word
        if n == 0:
            break

        # shortest prefix is entire list
        if n == len(visemes):
            return successes + [current + [visemes[0:n]]]

        # recurse over everything except this prefix
        successes += find_possible_chunks(visemes[n:], current + [visemes[:n]])
    return successes
\end{lstlisting}

\begin{lstlisting}[language=Python, caption={Find shortest prefix}\label{listing::find_shortest_prefix}]
def find_shortest_prefix(visemes, min_length=0):
    visemes = tuple(visemes)
    buffer = visemes[0:min_length+1]
    while not buffer in inverse.keys():
        if len(buffer)<len(visemes):
            buffer = visemes[0:len(buffer)+1]
            #prefix = len(buffer)
        else:
            prefix = 0
            return prefix
    prefix = len(buffer)
    return prefix
\end{lstlisting}

\begin{table}[!htbp]
\begin{center}
\caption{Word matches for a viseme sequence.}
\vspace{-0.6em}
\begin{tabular}{|c|c|}
\hline
\textbf{Viseme Cluster} & \textbf{Possible Words}\\ \hline
[w, ah, t, t] & \makecell{reits, reitz, rides, right's, rights, rights', rite's, rites, rust, ruts, rutts, rutz, what's,\\ whats, white's, whites, wide's, wised, wright's, wrights, writes}\\ \hline
[ah, p, iy, t] & abee, ip\\ \hline
[iy, t] & \makecell{tis, c's, c.'s, c.s, cease, cece, cede, cees, cid, cyd, d's, d.'s, d.s, deas, dease, dede,\\ dee's, deed, dees, deese, deet, deis, did, diede, dis, diss, dith, saez, scythe, sea's,\\ seas, sease, seat, seed, sees, seese, seethe, seis, seith, seize, sid, sies, siess, sis, sit,\\ syd, t's, t.'s, t.s, teas, tease, teat, teed, tees, teet, teeth, teethe, tese, thede, thee's,\\ these, thiede, thies, this, this', tidd, tiede, tis, tit, z's, z.'s, zeese, zeis}\\ \hline
\end{tabular}
\label{table:Word-matches-1}
\end{center}
\end{table}

\begin{table}[!htbp]
\begin{center}
\caption{Word matches for the next viseme sequence.}
\vspace{-0.6em}
\begin{tabular}{|c|c|}
\hline
\textbf{Viseme Cluster} & \textbf{Possible Words}\\ \hline
[w, ah, t, t] & \makecell{reits, reitz, rides, right's, rights, rights', rite's, rites, rust, ruts, rutts, rutz, what's,\\ whats, white's, whites, wide's, wised, wright's, wrights, writes}\\ \hline
[ah, p, iy] & abyss, amid, amiss, apiece, appease, eyepiece\\ \hline
[t, iy, t] & e's, e.'s, e.s, eade, ease, eat, ede, id, ihde, is, it\\ \hline
\end{tabular}
\label{table:Word-matches-2}
\end{center}
\end{table}

\subsection{Iterations for two scenarios}
\label{subsec:Iterations}

The viseme-to-word conversion process described in Subsection \ref{subsec:Word-Detector} is performed in an iterative process using a local beams search because an exhaustive search where the perplexity of all possible word combinations is calculated in either Scenarios 1 or 2 would not be feasible. For Scenario 1, cluster matching is only required for the one viseme sequence of clusters corresponding to the actual sentence, whereas for Scenario 2, the conversion process would be performed for all sequences of clusters outputted by the chunkification procedure described in Subsection \ref{subsec:Chunkification}.

The overall iterative procedure for Scenario 1 is illustrated in Figure \ref{fig::scenario-1}. If the sequence consists of only one cluster, then the most frequently appearing word according to the COCA word rankings is selected. If on the other hand the sequence consists of two or more clusters, perplexity iterations are then performed where the first pair of clusters are combined and matched to every set of possible words. The perplexity score of every combination of words is calculated with the minimum 50 being selected and combined with words corresponding to the next cluster. The iterative process is repeated until the end of the viseme sequence is reached. The sentence with the lowest perplexity score is selected as the final output.

\begin{figure}[!htbp]
\begin{tikzpicture}[node distance=2cm]
\node (start) [startstop] {Start};
\node (in1) [io, text width=8em, below = 0.8cm of start] {Sequences of Viseme Clusters};
\node (dec0) [decision, aspect=1, text width=7em, below = 0.8cm of in1] {Are there sequences with only 1 Cluster?};
\node (prof) [process, text width=10em, right = 3.58cm of dec0] {Select most frequent word for single word};
\node (proa) [process, text width=10em, below = 0.8cm of dec0] {Match first pair of viseme clusters to all possible Words};
\node (prob) [process, text width=10em, below = 0.8cm of proa] {Calculate perplexity};
\node (proc) [process, text width=10em, below = 0.8cm of prob] {Keep 50 combinations with minimum score};
\node (dec1) [decision, aspect=1, text width=8em, below = 0.8cm of proc, inner sep=0pt] {Has every\\ posssible viseme sequence reached an end?};
\node (procd) [process, text width=10em, below = 0.8cm of dec1] {Group sentences with words correpsonding to next cluster};
\node (proe) [process, text width=7em, right = 0.8cm of dec1] {Pick Minimum\\ Score};
\node (out1) [io, text width=4em, right = 0.4cm of proe] {Predicted Sentence};
\node (stop) [startstop, right = 0.4cm of out1] {End};
\node (coord) [coordinate, left = 2.0cm of proc] {}; 

\draw [arrow](start) -- (in1);
\draw [arrow](in1) -- (dec0);
\draw [arrow](dec0) -- node[anchor=west] {No} (proa);
\draw [arrow](proa) -- (prob);
\draw [arrow](prob) -- (proc);
\draw [arrow](proc) -- (dec1);
\draw [arrow](dec1) -- node[anchor=south] {Yes} (proe);
\draw [arrow](dec1) -- node[anchor=west] {No} (procd);
\draw [arrow](proe) -- (out1);
\draw [arrow](out1) -- (stop);
\draw [arrow](dec0) -- node[anchor=south] {Yes} (prof);
\draw [arrow](prof) -- (out1);
\draw [-](procd) -| (coord);
\draw [arrow](coord) |- (prob);

\end{tikzpicture}
\caption{Process map of viseme-to-word conversion for Scenario 2.}
\label{fig::scenario-2}
\end{figure}
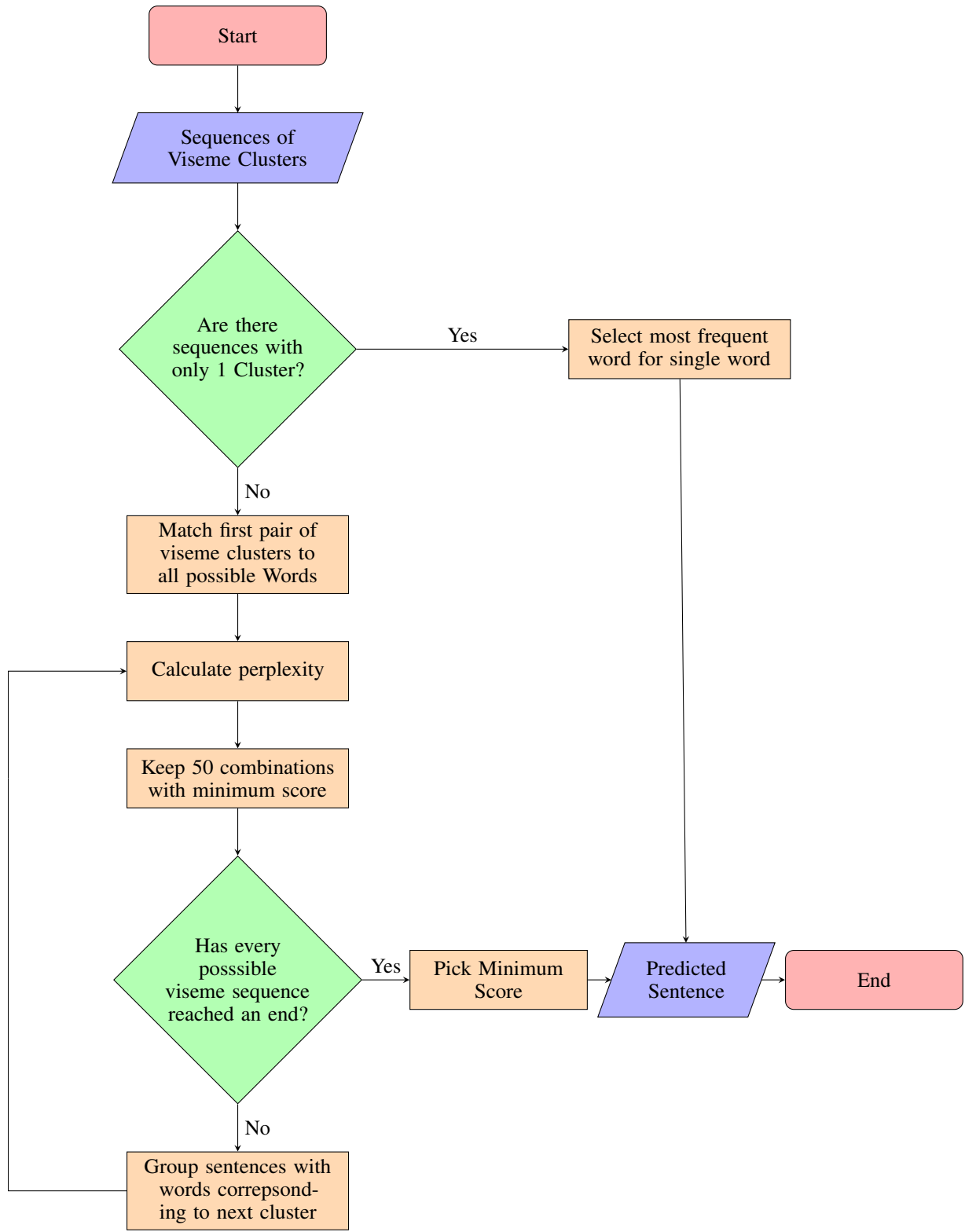

In the case of Scenario 2(illustrated in Figure \ref{fig::scenario-2}), if there are viseme sequences with only 1 cluster, then the most frequently appearing word that matches any of the individual clusters according to the COCA\cite{b29} word rankings is selected. If there are no sequences with single clusters, the first pair of viseme clusters for every viseme sequence gets matched to every possible combination of words contained within the lexicon and the perplexity score for every single combination is calculated. Many of the viseme sequences will share the same viseme clusters at the start of the sequences which helps reduce the number of operations needed for the matching process. 

The 50 combinations with minimum scores are stored and then the next cluster for every viseme sequence that begins with viseme clusters matching to the 50 phrases from the previous iteration gets matched to words from the lexicon and combined with those phrases, once again keeping only the 50 phrases with the lowest perplexity. This process continues iteratively until words from the longest viseme sequence are matched. However, if one of the stored phrases already maps to a full sequence of viseme clusters then no further combinations are made but the phrase will still be stored if it falls within phrases with the lowest 50 perplexity scores at any iteration. If at any stage of the iterations, all 50 sentences match entire viseme sequences in the input, the iterations are stopped and sentence with the lowest perplexity score is selected as the final output.

\begin{figure}[!htbp]
\begin{tikzpicture}[node distance=2cm]
\node (start) [startstop] {Start};
\node (in1) [io, text width=8em, below = 0.8cm of start] {Sequence of Viseme Clusters};
\node (dec0) [decision, aspect=2, below = 0.8cm of in1] {Only 1 Cluster?};
\node (prof) [process, text width=8em, right = 4.36cm of dec0] {Select most frequent word};
\node (proa) [process, text width=10em, below = 0.8cm of dec0] {Match Viseme\\ Clusters to Words};
\node (prob) [process, text width=10em, below = 0.8cm of proa] {Calculate perplexity\\ of word combinations};
\node (proc) [process, text width=10em, below = 0.8cm of prob] {Keep 50 combinations with minimum score};
\node (dec1) [decision, aspect=2, text width=7em, below = 0.8cm of proc] {End of Viseme sequence?};
\node (procd) [process, text width=10em, below = 0.8cm of dec1] {Combine sentences\\ with words matching to next cluster};
\node (proe) [process, text width=8em, right = 0.8cm of dec1] {Pick\\ Minimum Score};
\node (out1) [io, text width=5em, right = 0.4cm of proe] {Predicted Sentence};
\node (stop) [startstop, right = 0.4cm of out1] {End};
\node (coord) [coordinate, left = 1.2cm of proc] {}; 

\draw [arrow](start) -- (in1);
\draw [arrow](in1) -- (dec0);
\draw [arrow](dec0) -- node[anchor=west] {No} (proa);
\draw [arrow](proa) -- (prob);
\draw [arrow](prob) -- (proc);
\draw [arrow](proc) -- (dec1);
\draw [arrow](dec1) -- node[anchor=south] {Yes} (proe);
\draw [arrow](dec1) -- node[anchor=west] {No} (procd);
\draw [arrow](proe) -- (out1);
\draw [arrow](out1) -- (stop);
\draw [arrow](dec0) -- node[anchor=south] {Yes} (prof);
\draw [arrow](prof) -- (out1);
\draw [-](procd) -| (coord);
\draw [arrow](coord) |- (prob);

\end{tikzpicture}
\caption{Process map of viseme-to-word conversion for Scenario 1.}
\label{fig::scenario-1}
\end{figure}
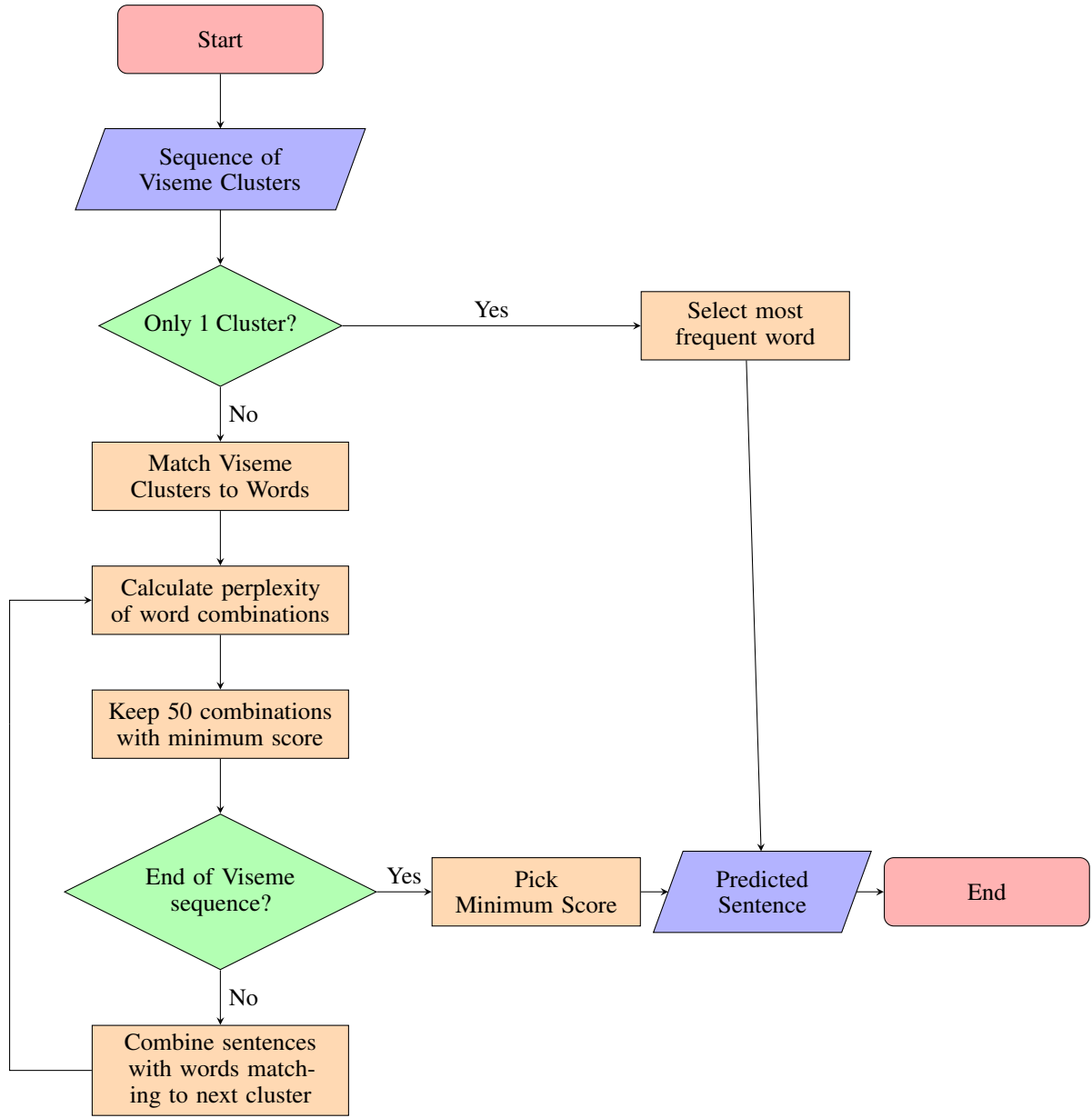

\newpage
\subsection{Systems Performance Measures}
\label{subsec:Measures}
The measures that have been used to evaluate the lip reading sentence system are edit distance-based metrics and are computed by calculating the normalized edit distance between the ground truth and a predicted sentence. Metrics reported in this paper include Viseme Error Rate (VER), Character Error Rate (CER), Word Error Rates (WER) and Sentence Accuracy Rate (SAR).

Error rate metrics used for evaluating accuracy are given by calculating the overall edit distance. In determining misclassifications, one has to compare the decoded speech to the actual speech. The equation for calculating Error Rate (ER) is given in Eq. \ref{ER} with $N$ being the total number of characters in the ground truth, $S$ being the number of characters substituted for wrong classifications, $I$ being the number of characters inserted for those not picked up and $D$ being the number of deletions being made for decoded characters that should not be present. CER, WER and VER are all calculated this way with the expressions given in Eqs. \ref{CER}, \ref{WER} and \ref{VER} where $C$, $W$ and $V$ correspond to characters, words and visemes.

\vspace{-0.8em}
\begin{equation} ER=\frac{S+D+I}{N} \label{ER}\end{equation}
\begin{equation} CER=\frac{C_{S}+C_{D}+C_{I}}{C_{N}} \label{CER}\end{equation}
\begin{equation} WER=\frac{W_{S}+W_{D}+W_{I}}{W_{N}} \label{WER}\end{equation}
\begin{equation} VER=\frac{V_{S}+V_{D}+V_{I}}{V_{N}} \label{VER}\end{equation}

SAR is a binary metric as expressed in Eq. \ref{SAR}, where the value is 1 if the predicted sentence $P_P$ is equal to the ground truth $P_T$, otherwise it would take the value of 0:

\vspace{-0.8em}
\begin{equation}SAR=\begin{cases}
					1, & P_P=P_T\\
					0, & P_P \ne P_T
					\end{cases}
\label{SAR}
\end{equation}

\section{Experiment and Results}
\label{sec:Experiment}

This section consists of explanatory details of the dataset used to provide the viseme clusters that are used as samples to evaluate the accuracy of the word detector, and a subsection giving accuracy results of the word detector for both Scenarios 1 and 2 which discussions on how the performance can be improved.

\subsection{Datasets}
\label{subsec:Datasets}

This paper has utilised two datasets: the OuluVS\cite{b40} dataset consisting of short phrases and the BBC LRS2\cite{b6} dataset consisting of longer phrases. The OuluVS corpus has a set of ten phrases including "hello", "excuse me", "I am sorry", "thank you", "good bye", "I love this game", "nice to meet you", "you are welcome", "how are you" and "have a good time" whereas the BBC LRS2 dataset has longer and more random sentences. 

The LRS2 dataset consists sentences of up to 100 characters from BBC videos and is extremely challenging due to the variety in genres and the number of speakers. The "test" section of the corpus contains 1243 sentences with 6660 word instances and a vocabulary range of 1697 possible words. A beam width of 50 has been use for both Scenarios 1 and 2.

\subsection{Results}
\label{subsec:Results}

Table \ref{table:metric-results} shows the overall accuracy metrics for the optimum sentences predicted for both Scenarios 1 and 2. Table \ref{table:predicted-sentences} gives an example of some of the sentences that were chosen in either Scenario 1 or 2 as well as their perplexity scores according to the GPT.

\begin{table}[!htbp]
\begin{center}
\caption{Metrics for predicted sentence results.}
\vspace{-0.6em}
\begin{tabular}{|c|c|c|c|c|c|c|c|}
\hline
\multirow{2}{*}{\textbf{Dataset}} & \multirow{2}{*}{\textbf{Number of Sentences}} & \multicolumn{3}{c|}{\textbf{Scenario 1}} & \multicolumn{3}{c|}{\textbf{Scenario 2}}\\ \cline{3-8}
& & \textbf{CER(\%)} & \textbf{WER(\%)} & \textbf{SAR(\%)} & \textbf{CER(\%)} & \textbf{WER(\%)} & \textbf{SAR(\%)}\\ \hline
\textbf{OuluVS} & 10 & 0.0 & 0.0 & 100.0 & 0.0 & 0.0 & 100.0\\ \hline
\textbf{LRS} & 1243 & 10.7 & 18.0 & 56.8 & 36.1 & 48.3 & 35.1\\ \hline
\end{tabular}
\label{table:metric-results}
\end{center}
\end{table}

\begin{table}[!htbp]
\begin{scriptsize}
\begin{center}
\caption{Predicted sentence examples.}
\vspace{-1.4em}
\begin{tabular}{|c|c|c|c|c|c|}
\hline
\multicolumn{2}{|c|}{\textbf{Actual Labels}} & \multicolumn{2}{c|}{\textbf{Scenario 1}} & \multicolumn{2}{c|}{\textbf{Scenario 2}}\\ \hline
\textbf{Actual Sentence} & \textbf{Perplexity} & \textbf{Predicted Sentence} & \textbf{Perplexity} & \textbf{Predicted Sentence} & \textbf{Perplexity}\\ \hline
\textbf{EXCUSE ME} & 5.3 & EXCUSE ME & 5.3 & EXCUSE ME & 5.3\\ \hline
\textbf{I AM SORRY} & 53.6 & I AM SORRY & 53.6 & I AM SORRY & 53.6\\ \hline
\textbf{I LOVE THIS GAME} & 94.0 & I LOVE THIS GAME & 94.0 & I LOVE THIS GAME & 94.0\\ \hline
\makecell{\textbf{WHEN THERE ISN'T MUCH} \\\textbf{ELSE IN THE GARDEN}} & 60.3 & \makecell{WHEN THERE ISN'T MUCH \\ELSE IN THE GARDEN} & 60.3 & \makecell{WHEN THERE ISN'T MUCH \\ELSE IN THE GARDEN} & 60.3\\ \hline
\textbf{OVER THE COURSE OF HIS LIFE} & 7.8 & OVER THE COURSE OF HIS LIFE & 7.8 & OVER THE COURSE OF HIS LIFE & 7.8\\ \hline
\makecell{\textbf{BUT NOW WE} \\\textbf{HAVE THESE VIRUSES}} & 285.4 & \makecell{BUT NOW WE \\HAVE THIS VIRUS} & 103.3 & \makecell{BUT NOW WE \\HAVE THESE VIRUSES} & 285.4\\ \hline
\textbf{PRETTY ON THE OUTSIDE} & 75.5 & BREEZY ON THE OUTSIDE & 61.2 & BREEZY ON THE OUTSIDE & 61.2\\ \hline
\makecell{\textbf{STICK TO WHAT} \\\textbf{YOU'RE GOOD AT}} & 81.1 & \makecell{STILL DO WHAT \\YOU'RE GOOD AT} & 86.9 & \makecell{STILL DO WHAT \\YOU'RE GOOD AT} & 86.9\\ \hline
\end{tabular}
\label{table:predicted-sentences}
\end{center}
\end{scriptsize}
\end{table}

For all ten phrases of the OuluVS corpus; the correct phrase was chosen – including Scenario 1 which is important given that the segmentation of words using only visual information remains an ongoing challenge. For the more challenging LRS2 corpus, words were predicted correctly on the majority of occasions where visemes were segmented as can be seen by looking at Scenario 1 performance results. The performance for Scenario 2 is not as good and this is expected as in many circumstances, the boundaries between words were not even correctly recognised.

Aside from needing to segment viseme clusters, the classification performance has also been hindered in some circumstances by the presence of local optima in the beam search. The sentence "stick to what you're good it" was predicted as "still do what you're good at" despite the fact that the predicted sentence has a higher perplexity score than the actual sentence. However, the bigram "still do" has a significantly lower perplexity than "stick to" and the latter bigram does not even fall into the list of bigrams with the minimum 50 perplexities during the word matching process.

The performance results could significantly be improved further by increasing the local beam width, though this does come at the cost of increasing the computational overhead required. There are alternative heuristic search algorithms that could be used instead of local beam search and further optimisation can be done to enhance the accuracy of viseme-to-word conversion and to tackle the problem of local optima where at some iterations the ground truth is not contained within the top 50 results.

\section{Conclusion}
\label{sec:Conclusion}

It has been demonstrated that through using a viseme-based classification base lip reading system,  the speech of an individual can be successfully predicted given the challenge one-to-many mapping between visemes and phonemes and the presence of homopheme words and for all ten phrases were predicted correctly. Given that many languages share common visemes, one could hypothetically use perplexity score given by a transformer trained in another language and apply a viseme-based lip reading system to someone speaking in that other language.

The overall accuracy of the conversion of visemes to words can be further improved upon by experimenting with different global search optimisation methods while also limiting the computational overhead required. Furthermore, the model should be adapted to deal with circumstances when visemes are misclassified.

%\bibliographystyle{unsrtnat}
%\bibliography{references}  %%% Uncomment this line and comment out the ``thebibliography'' section below to use the external .bib file (using bibtex) .

 Uncomment this section and comment out the \bibliography{references} line above to use inline references.

\end{document}